\documentclass[11pt,a4paper]{article}

\usepackage{acl2015}
\usepackage{times}
\usepackage{latexsym}
\usepackage{amsmath}
\usepackage{multirow}
\usepackage{url}

\usepackage{rotating}

\usepackage{xcolor}

\usepackage{qtree}

\definecolor{myblue}{rgb}{0.25,0.41,0.88}
\definecolor{mygreen}{rgb}{0.15,0.55,0.11}
\definecolor{myred}{rgb}{0.88,0.41,0.25}

\newcommand{\e}[1]{\hat{#1}}

\usepackage[noend]{algorithmic}
\usepackage{algorithm}
\usepackage{xspace}
\usepackage{amsmath}

\newfont{\msym}{msbm10}
\newcommand{\reals}{\mbox{\msym R}}

\newcommand{\xione}{t^{(i)}}

\newcommand{\aione}{a^{(i)}}

\newcommand{\yione}{y^{(i)}}

\newcommand{\bi}{z^{(i)}}
\newcommand{\oi}{o^{(i)}}
\newcommand{\myxi}{x^{(i)}}

\newcommand{\p}{{\cal P}}
\newcommand{\internal}{{\cal I}}
\newcommand{\n}{{\cal N}}

\newcommand{\myepsilon}{\varepsilon}

\newcommand{\setupAll}{{\sf All}}
\newcommand{\setupDropout}{{\sf Dropout}}
\newcommand{\setupAdd}{{\sf Add}}
\newcommand{\setupMul}{{\sf Mul}}

\title{Diversity in Spectral Learning for Natural Language Parsing}
\author{
Shashi Narayan and Shay B. Cohen \\ School of Informatics \\ University of Edinburgh \\ Edinburgh, EH8 9LE, UK \\ \texttt{\{snaraya2,scohen\}@inf.ed.ac.uk}
}

\makeatletter
\newcommand{\@BIBLABEL}{\@emptybiblabel}
\newcommand{\@emptybiblabel}[1]{}
\makeatother

\begin{document}

\maketitle

\begin{abstract}
  We describe an approach to create a diverse set of predictions with spectral
  learning of latent-variable PCFGs (L-PCFGs). Our approach works by
  creating multiple spectral models where noise is added to the
  underlying features in the training set before the estimation of
  each model. We describe three ways to decode with multiple
  models. In addition, we describe a simple variant of the spectral
  algorithm for L-PCFGs that is fast and leads to compact models. Our
  experiments for natural language parsing, for English and German,
  show that we get a significant improvement over baselines comparable
  to state of the art. For English, we achieve the $F_1$ score of
  90.18, and for German we achieve the $F_1$ score of 83.38.
\end{abstract}

\section{Introduction}


It has been long identified in NLP that a diverse set of solutions
from a decoder can be reranked or recombined in order to improve the
accuracy in various problems \cite{henderson1999exploiting}. Such
problems include machine translation \cite{macherey2007empirical},
syntactic parsing
\cite{charniak-05,sagae2006parser,fossum2009combining,zhang2009k,petrov2010products,choe-15}
and others \cite{van2001improving}.

The main argument behind the use of such a diverse set of solutions
(such as $k$-best list of parses for a natural language sentence) is
the hope that each solution in the set is mostly correct.  Therefore,
recombination or reranking of solutions in that set will further
optimize the choice of a solution, combining together the information
from all solutions.

In this paper, we explore another angle for the use of a set of parse
tree predictions, where all predictions are made for the same
sentence.  More specifically, we describe techniques to exploit
diversity with spectral learning algorithms for natural language
parsing. Spectral techniques and the method of moments have been
recently used for various problems in natural language processing,
including parsing, topic modeling and the derivation of word
embeddings \cite{luque-12,cohen-13b,stratos2014spectral,dhillon-15,rastogimultiview,nguyen2015your,lu-15}.

\newcite{cohen-13b} showed how to estimate an L-PCFG using spectral
techniques, and showed that such estimation outperforms the
expectation-maximization algorithm \cite{matsuzaki-05}. Their result
still lags behind state of the art in natural language parsing, with
methods such as coarse-to-fine \cite{petrov-06}.

We further advance the accuracy of natural language parsing with
spectral techniques and L-PCFGs, yielding a result that outperforms
the original Berkeley parser from \newcite{petrov-07}.  Instead of
exploiting diversity from a $k$-best list from a single model, we
estimate multiple models, where the underlying features are perturbed
with several perturbation schemes. Each such model, during test time,
yields a single parse, and all parses are then used together in
several ways to select a single best parse.

The main contributions of this paper are two-fold. First, we present
an algorithm for estimating L-PCFGs, akin to the spectral algorithm of
\newcite{cohenacl2012}, but simpler to understand and implement. This
algorithm has value for readers who are interested in learning
more about spectral algorithms -- it demonstrates some of the core
ideas in spectral learning in a rather intuitive way.
In addition, this algorithm leads to sparse grammar estimates and compact
models.

Second, we describe how a diverse set of predictors can be used with
spectral learning techniques. Our approach relies on adding noise to
the feature functions that help the spectral algorithm compute the
latent states. Our noise schemes are similar to those described by
\newcite{wang2013feature}. We add noise to the whole
training data, then train a model using our algorithm (or other spectral
algorithms; Cohen et al., 2013\nocite{cohen-13b}), and repeat this
process multiple times. We then use the set of parses we get from
all models in a recombination step.





The rest of the paper is organized as follows. In
\S\ref{section:background} we describe notation and background about
L-PCFG parsing.  In \S\ref{section:alg} we describe our new spectral
algorithm for estimating L-PCFGs. It is based on similar intuitions as
older spectral algorithms for L-PCFGs. In \S\ref{section:noise} we
describe the various noise schemes we use with our spectral algorithm
and the spectral algorithm of \newcite{cohen-13b}. In
\S\ref{section:decoding} we describe how to decode with multiple
models, each arising from a different noise setting. In
\S\ref{section:experiments} we describe our experiments with natural
language parsing for English and German.







\section{Background and Notation}
\label{section:background}

We denote by $[n]$ the set of integers $\{ 1, \ldots, n \}$. For
a statement $\Gamma$, we denote by $[[ \Gamma ]]$ its indicator
function, with values 0 when the assertion is false and 1 when it is true.

An L-PCFG is a 5-tuple $(\n, \internal, \p, m, n)$ where:

\begin{itemize}

\item $\n$ is the set of nonterminal symbols in the grammar.
  $\internal \subset \n$ is a finite set of {\em interminals}.  $\p
  \subset \n$ is a finite set of {\em preterminals}.  We assume that
  $\n = \internal \cup \p$, and $\internal \cap \p= \emptyset$.  Hence
  we have partitioned the set of nonterminals into two subsets.

\item $[m]$ is the set of possible hidden states.

\item $[n]$ is the set of possible words.

\item For all $a \in \internal$, $b \in \n$, $c \in \n$,
  $h_1, h_2, h_3 \in [m]$, we have a binary context-free rule
$
a(h_1) \rightarrow b(h_2) \;\; c(h_3)
$.

\item For all $a \in \p$, $h \in [m]$, $x \in [n]$, we have a lexical
  context-free rule $ a(h) \rightarrow x $.

\end{itemize}

Latent-variable PCFGs are essentially equivalent to probabilistic
regular tree grammars (PRTGs; Knight and Graehl,
2005\nocite{knight2005overview}) where the righthand side trees are of
depth 1. With general PRTGs, the righthand side can be
of arbitrary depth, where the leaf nodes of these trees correspond to
latent states in the L-PCFG formulation above and the internal nodes
of these trees correspond to interminal symbols in the L-PCFG
formulation.

\begin{figure}
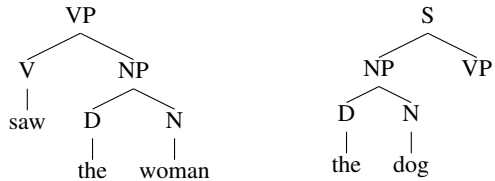

\begin{center}
\begin{footnotesize}
\begin{tabular}{lp{0.3in}l}
\Tree [.VP [.V saw ] [.NP [.D the ] [.N woman ] ] ]
&
&
\Tree [.S [.NP [.D the ] [.N dog ] ] VP ]
\end{tabular}
\end{footnotesize}
\end{center}
\caption{The inside tree (left) and outside tree (right)
  for the nonterminal {\tt VP} in the parse tree
{\tt (S (NP (D the) (N dog)) (VP (V saw) (NP (D the) (N woman)))).}\vspace{-2ex}}
\label{fig:iotrees}
\end{figure}

Two important concepts that will be used throughout of the paper are
that of an ``inside tree'' and an ``outside tree.''  Given a tree, the
inside tree for a node contains the entire subtree below that node;
the outside tree contains everything in the tree excluding the inside
tree. See Figure~\ref{fig:iotrees} for an example. Given a grammar, we
denote the space of inside trees by $T$ and the space of outside trees
by $O$.

\begin{figure}[t]
\begin{footnotesize}
\framebox{\parbox{\columnwidth}{ 

{\bf Inputs:} An input treebank with the following additional information: training examples 
$(\aione, \xione, \oi, b^{(i)})$
for $i \in \{1 \ldots M\}$, where $\aione \in \n$; 
$\xione$ is an inside tree; 
$\oi$ is an outside tree; and $b^{(i)} = 1$ if the rule is at the root
of tree, $0$ otherwise.
A function $\phi$ that maps inside trees $t$
to feature-vectors $\phi(t) \in \reals^d$. A function $\psi$ that
maps outside trees $o$ to feature-vectors $\psi(o) \in \reals^{d'}$.
An integer $k$ denoting the thin-SVD rank. 
An integer $m$ denoting the number of latent states.

$\,$

{\bf Algorithm:}

(Step 1: Singular Value Decompositions)

\begin{itemize}

\item Calculate SVD on $\Omega^a$ to get
$\e{U}^a \in \reals^{(d \times k)}$ and $\e{V}^a
\in \reals^{(d' \times k)}$ for each $a \in \n$.

\end{itemize}

(Step 1: Projection)

\begin{itemize}

\item For all $i \in [M]$, compute $\yione = (\e{U}^{a_i})^\top \phi(\xione)$ and $\bi = (\e{V}^{a_i})^\top \psi(\oi)$.

\item For all $i \in [M]$, set $\myxi$ to be the concatenation of $\yione$ and $\bi$.

\end{itemize}

(Step 2: Cluster Projections)

\begin{itemize}

\item For all $a \in \n$, cluster the set $\{ \myxi \mid \aione = a \}$ to get a
clustering function $\gamma \colon \reals^{2k} \rightarrow [m]$ that maps a projected
vector $\myxi$ to a cluster in $[m]$.

\end{itemize}

(Step 3: Compute Final Parameters)

\begin{itemize}

\item Annotate each node in the treebank with $\gamma(\myxi)$.

\item Compute the probability of a rule $p(a[h_1] \rightarrow b[h_2] \, c[h_3] \mid a[h_1])$ as
the relative frequency of its appearance in the cluster-annotated treebank.

\item Similarly, compute the root probabilities $\pi(a[h])$ and preterminal
rules $p(a[h] \rightarrow x \mid a[h])$.

\end{itemize}

}}

\end{footnotesize}
\caption{The clustering estimation algorithm for L-PCFGs.}
\label{fig:mlelearn}

\end{figure}

\section{Clustering Algorithm for Estimating L-PCFGs}
\label{section:alg}


We assume two feature functions, $\phi \colon T \rightarrow \reals^d$
and $\psi \colon O \rightarrow \reals^{d'}$, mapping inside and
outside trees, respectively, to a real vector. Our training data
consist of examples $(\aione, \xione, \oi, b^{(i)})$ for $i \in \{1
\ldots M\}$, where $\aione \in \n$; $\xione$ is an inside tree; $\oi$
is an outside tree; and $b^{(i)} = 1$ if $\aione$ is the root of tree,
$0$ otherwise. These are obtained by splitting all trees in the
training set into inside and outside trees at each node in each tree.
We then define $\Omega^a \in \reals^{d \times d'}$:

\begin{equation}
\Omega^a = \frac{\sum_{i=1}^M [[\aione = a ]] \phi(\xione) (\psi(\oi))^\top}
  {\sum_{i=1}^M [[\aione =a ]]}
\end{equation}

This matrix is an empirical estimate for the
cross-covariance matrix between the inside trees and the outside trees
of a given nonterminal $a$.  An inside tree and an outside tree are
conditionally independent according to the L-PCFG model, when the
latent state at their connecting point is known. This means that the
latent state can be identified by finding patterns that co-occur
together in inside and outside trees -- it is the only random variable
that can explain such correlations.  As such, if we reduce the
dimensions of $\Omega^a$ using singular value decomposition (SVD), we
essentially get representations for the inside trees and the outside
trees that correspond to the latent states.

This intuition leads to the algorithm that appears in
Figure~\ref{fig:mlelearn}.  The algorithm we describe takes as input training
data, in the form of a treebank, decomposed into inside and outside
trees at each node in each tree in the training set.

The algorithm first performs SVD for each of the set of inside and
outside trees for all nonterminals.\footnote{We normalize features
  by their variance.} This step is akin to CCA, which has been used in
various contexts in NLP, mostly to derive representations for words
\cite{dhillon-15,rastogimultiview}.  The algorithm then takes the
representations induced by the SVD step, and clusters them -- we use
$k$-means to do the clustering. Finally, it maps each SVD
representation to a cluster, and as a result, gets a cluster
identifier for each node in each tree in the training data. These
clusters are now treated as latent states that are ``observed.'' We
subsequently follow up with frequency count maximum likelihood
estimate to estimate the probabilities of each parameter in the
L-PCFG.

Consider for example the estimation of rules of the form $a \rightarrow x$.
Following the clustering step we obtain for each nonterminal $a$ and latent state $h$ a set of
rules of the form $a[h] \rightarrow x$. Each such instance comes from
a single training example of a lexical rule.  Next, we compute the
probability of the rule $a[h] \rightarrow x$ by counting how many times
that rule appears in the training instances, and normalize by the
total count of $a[h]$ in the training instances. Similarly, we compute
probabilities for binary rules of the form $a \rightarrow b \, c$.

The features that we use for $\phi$ and $\psi$ are similar to those
used in \newcite{cohen-13b}.  These features look at the local
neighborhood surrounding a given node. More specifically, we indicate
the following information with the inside features (throughout these
definitions assume that $a \rightarrow b \, c$ is at the root of the
inside tree $t$):

\begin{itemize}\setlength\itemsep{-0.2em}
\item The pair of nonterminals $(a, b)$. E.g., for the inside tree in
  Figure~\ref{fig:iotrees} this would be the pair (VP, V).
\item The pair $(a, c)$. E.g., (VP, NP).
\item The rule $a \rightarrow b \, c$. E.g., VP $\rightarrow$ V NP.
\item The rule $a \rightarrow b \, c$ paired with the rule at the node
  $b$. E.g., for the inside tree in Figure~\ref{fig:iotrees} this
  would correspond to the tree fragment (VP (V saw) NP).
\item The rule $a \rightarrow b \, c$ paired with the rule at the node
  $c$. E.g., the tree fragment (VP V (NP D N)).
\item The head part-of-speech of $t$ paired with $a$. E.g., the pair
  (VP, V).
\item The number of words dominated by $t$ paired with $a$. E.g., the
  pair (VP, 3).
\end{itemize}

In the case of an inside tree consisting of a single rule $a
\rightarrow x$ the feature vector simply indicates the identity of
that rule.

For the outside features, we use:

\begin{itemize}\setlength\itemsep{-0.2em}
\item The rule above the foot node.  E.g., for the outside tree in
  Figure~\ref{fig:iotrees} this would be the rule S $\rightarrow$ NP
  VP$^*$ (the foot nonterminal is marked with $*$).
\item The two-level and three-level rule fragments above the foot
  node. These features are absent in the outside tree in
  Figure~\ref{fig:iotrees}.
\item The label of the foot node, together with the label of its
  parent. E.g., the pair (VP, S).
\item The label of the foot node, together with the label of its
  parent and grandparent. 
\item The part-of-speech of the first head word along the path from
  the foot of the outside tree to the root of the tree which is
  different from the head node of the foot node.
\item The width of the spans to the left and to the right of the foot
  node, paired with the label of the foot node.
\end{itemize}

\paragraph{Other Spectral Algorithms} The SVD step on the $\Omega^a$
matrix is pivotal to many algorithms, and has been used in the past
for other L-PCFG estimation algorithms. \newcite{cohenacl2012} used it
for developing a spectral algorithm that identifies the parameters of
the L-PCFG up to a linear transformation. Their algorithm generalizes
the work of \newcite{hsu09} and \newcite{bailly-10}.

\newcite{cohen-14b} also developed an algorithm that makes use of an
SVD step on the inside-outside. It relies on the idea of ``pivot
features'' -- features that uniquely identify latent states.

\newcite{louis-15} used a clustering algorithm that resembles ours but
does not separate inside trees from outside trees or follows up with
a singular value decomposition step. Their algorithm was applied
to both L-PCFGs and linear context-free rewriting systems. Their application
was the analysis of hierarchical structure of conversations in online forums.


In our preliminary experiments, we found out that the clustering
algorithm by itself performs worse than the spectral algorithm of
\newcite{cohen-13b}. We believe that the reason is two-fold: (a)
$k$-means finds a local maximum during clustering; (b) we do hard
clustering instead of soft clustering. However, we detected that the
clustering algorithm gives a more diverse set of solutions, when the
features are perturbed. As such, in the next sections, we explain how
to perturb the models we get from the clustering algorithm (and the
spectral algorithm) in order to improve the accuracy of the clustering
and spectral algorithms.

\section{Spectral Estimation with Noise}
\label{section:noise}


It has been shown that a diverse set of predictions can be used to help improve decoder accuracy
for various problems in NLP \cite{henderson1999exploiting}. Usually a
$k$-best list from a single model is used to exploit model
diversity. Instead, we estimate multiple models, where the underlying
features are filtered with various noising schemes.

We try three different types of noise schemes for the algorithm in
Figure~\ref{fig:mlelearn}:

\begin{description}
\item[Dropout noise:] Let $\sigma \in [0,1]$. We set each
  element in the feature vectors $\phi(t)$ and $\psi(o)$ to $0$ with
  probability $\sigma$.
\item[Gaussian (additive):] Let $\sigma > 0$. For each
  $x^{(i)}$, we draw a vector $\myepsilon \in \reals^{2k}$ of
  Gaussians with mean 0 and variance $\sigma^2$, and then set $x^{(i)}
  \leftarrow x^{(i)} + \myepsilon$.
\item[Gaussian (multiplicative):] Let $\sigma > 0$. For each
  $x^{(i)}$, we draw a vector $\myepsilon \in \reals^{2k}$ of
  Gaussians with mean 0 and variance $\sigma^2$, and then set $x^{(i)}
  \leftarrow x^{(i)} \otimes (1 + \myepsilon)$, where $\otimes$ is
  coordinate-wise multiplication.
\end{description}

Note the distinction between the dropout noise and the Gaussian noise
schemes: the first is performed on the feature vectors before the SVD
step, and the second is performed after the SVD step. It is not
feasible to add Gaussian noise prior to the SVD step, since the matrix
$\Omega^a$ will no longer be sparse, and its SVD computation will be
computationally demanding.

Our use of dropout noise here is inspired by ``dropout'' as is used in
neural network training, where various connections between units in
the neural network are dropped during training in order to avoid
overfitting of these units to the data \cite{srivastava2014dropout}.

The three schemes we described were also used by
\newcite{wang2013feature} to train log-linear models. Wang et al.'s
goal was to prevent overfitting by introducing this noise schemes as
additional regularizer terms, but without explicitly changing the
training data. We do filter the data through these noise schemes, and
show in \S\ref{section:experiments} that all of these noise schemes do
not improve the performance of our estimation on their own. However,
when multiple models are created with these noise schemes, and then
combined together, we get an improved performance. As such, our
approach is related to the one of \newcite{petrov2010products}, who
builds a committee of latent-variable PCFGs in order to improve a
natural language parser.

We also use these perturbation schemes to create multiple models for the
algorithm of \newcite{cohenacl2012}. The dropout scheme stays the same,
but for the Gaussian noising schemes, we follow a slightly different
procedure. After noising the projections of the inside and outside
feature functions we get from the SVD step, we use these
projected noised features as a new set of inside and outside feature
functions, and re-run the spectral algorithm of \newcite{cohenacl2012}
on them.

We are required to add this extra SVD step because the spectral algorithm of Cohen et
al. assumes the existence of linearly transformed parameter estimates,
where the parameters of each nonterminal $a$ is linearly transformed
by unknown invertible matrices.  These matrices cancel out when the
inside-outside algorithm is run with the spectral estimate output. In
order to ensure that these matrices still exactly cancel out, we have
to follow with another SVD step as described above. The latter SVD step
is performed on a dense $\Omega^a \in \reals^{m \times m}$ but this is not an
issue considering $m$ (the number of latent states) is much smaller
than $d$ or $d'$.


\section{Decoding with Multiple Models}
\label{section:decoding}

Let $G_1,\ldots,G_p$ be a set of L-PCFG grammars. In
\S\ref{section:experiments}, we create such models using the noising
techniques described above. The question that remains is how to
combine these models together to get a single best output parse tree
given an input sentence.


With L-PCFGs, decoding a single sentence requires marginalizing out
the latent states to find the best skeletal tree\footnote{A skeletal tree is a derivation tree without latent states decorating the nonterminals.}
for a given string.
Let $s$ be a sentence. We define $t(G_i, s)$ to be the output tree
according to minimum Bayes risk decoding.  This means
we follow \newcite{goodman-96}, who uses dynamic programming
to compute the tree that maximizes the sum of all marginals of all
nonterminals in the output tree. Each marginal, for each span $\langle
a,i,j \rangle$ (where $a$ is a nonterminal and $i$ and $j$ are
endpoints in the sentence), is computed by using the inside-outside
algorithm.

In addition, let $\mu(a,i,j | G_k, s)$ be the marginal, as computed by
the inside-outside algorithm, for the span $\langle a,i,j \rangle$
with grammar $G_k$ for string $s$. We use the notation $\langle a,i,j
\rangle \in t$ to denote that a span $\langle a,i,j \rangle$ is in a
tree $t$.

We suggest the following three ways for decoding with multiple models
$G_1,\ldots,G_p$:


\begin{description}

\item[Maximal tree coverage:] Using dynamic programming, we return the
  tree that is the solution to:

\vspace{-0.5cm}
\begin{equation*}
  t^{\ast} = \arg\max_t \sum_{\langle a,i,j \rangle \in t} \sum_{k=1}^p [[ \langle a,i,j\rangle \in t(G_k, s)]].
\end{equation*}

This implies that we find the tree that maximizes its coverage with
respect to all other trees that are decoded using $G_1,\ldots,G_p$.

\item[Maximal marginal coverage:] Using dynamic programming, we return
  the tree that is the solution to:

\vspace{-0.5cm}
\begin{equation*}
  t^{\ast} = \arg\max_t \sum_{\langle a,i,j \rangle \in t} \sum_{k=1}^p \mu(a,i,j | G_k, s).
\end{equation*}

This is similar to maximal tree coverage, only instead of considering
just the single decoded tree for each model among $G_1,\ldots,G_p$, we
make our decoding ``softer,'' and rely on the marginals that each
model gives.

\item[MaxEnt reranking:] We train a MaxEnt reranker on a training set
  that includes outputs from multiple models, and then, during testing
  time, decode with each of the models, and use the trained reranker
  to select one of the parses.  We use the reranker of
  \newcite{charniak-05}.\footnote{Implementation:
    \url{https://github.com/BLLIP/bllip-parser}. More specifically, we
    used the programs {\tt extract-spfeatures}, {\tt cvlm-lbfgs} and
    {\tt best-indices}.  {\tt cvlm-lbfgs} was used with the default
    hyperparameters from the Makefile.}
\end{description}

As we see later in \S\ref{section:experiments}, it is sometimes
possible to extract more information from the training data by using a
network, or a hierarchy of the above tree combination methods. For
example, we get our best result for parsing by first using MaxEnt with
several subsets of the models, and then combining the output of these
MaxEnt models using maximal tree coverage.

\section{Experiments}
\label{section:experiments}

In this section, we describe parsing experiments with two languages:
English and German.

\subsection{Results for English}




\begin{table*}[htb]
  \begin{center}
    {\small
      \begin{tabular}{|c||c|c|c||c|c|c||c|c|c|}
        \hline
        & \multicolumn{3}{c||}{Clustering} & \multicolumn{3}{c||}{Spectral (smoothing)} & \multicolumn{3}{c|}{Spectral (no smoothing)} \\
        & MaxTre & MaxMrg & MaxEnt & MaxTre & MaxMrg & MaxEnt & MaxTre & MaxMrg & MaxEnt \\
        \hline
        \setupAdd & 88.68 & 88.64 & 89.50 & 88.20 & 88.28 & 88.59 & 86.72 & 86.85 & 87.94 \\
        \setupMul & 88.74 & 88.66 & 89.89  & 88.48 & 88.70 & 89.46 & 86.97 & 86.53 & 89.04 \\
        \setupDropout & 88.68 & 88.56 & 89.80 & 88.64 & 88.71 & \textbf{89.47} & 88.37 & 88.06 & 89.52 \\
        \setupAll & 88.84 & 88.75 & \textbf{89.95} & 88.38 & 88.75 & 89.45 & 87.49 & 87.00 & \textbf{89.85} \\
        \hline
        No noise & \multicolumn{3}{c||}{86.48} & \multicolumn{3}{c||}{88.53 (Cohen et al., 2013)} & \multicolumn{3}{c|}{86.47 (Cohen et al., 2013)} \\
        \hline
      \end{tabular}
    }
  \end{center}
  \caption{\small Results on section 22 (WSJ). MaxTre denotes decoding
    using maximal tree coverage, MaxMrg denotes decoding using maximal
    marginal coverage, and MaxEnt denotes the use of a discriminative
    reranker.  \setupAdd, \setupMul $\,$ and \setupDropout $\,$ denote
    the use of additive Gaussian noise, multiplicative Gaussian noise
    and dropout noise, respectively.  The number of models used in the
    first three rows for the clustering algorithm is 80: 20 for each
    $\sigma \in \{ 0.05, 0.1, 0.15, 0.2 \}$. For the spectral
    algorithm, it is 20, 5 for each $\sigma$ (see footnotes).  The
    number of latent states is $m=24$. For \setupAll, we use all
    models combined from the first three rows.  The ``No noise''
    baseline for the spectral algorithm is taken from Cohen et
    al. (2013). The best figure in each algorithm block is in
    boldface.\label{table:results-dev}}
\end{table*}

For our English parsing experiments, we use a standard setup.
More specifically, we use the Penn WSJ treebank \cite{marcus-93} for
our experiments, with sections 2--21 as the training data, and section
22 used as the development data. Section 23 is used as the final test
set. We binarize the trees in training data, but transform them back before
evaluating them.

For efficiency, we use a base PCFG without latent states to prune
marginals which receive a value less than $0.00005$ in the dynamic
programming chart.  The parser takes part-of-speech tagged sentences
as input. We tag all datasets using Turbo Tagger \cite{martins-10},
trained on sections 2--21.  We use the $F_1$ measure according to the
PARSEVAL metric \cite{black-91} for the evaluation.

\begin{figure}[htb]
  \centering
  \includegraphics[width=0.53\textwidth]{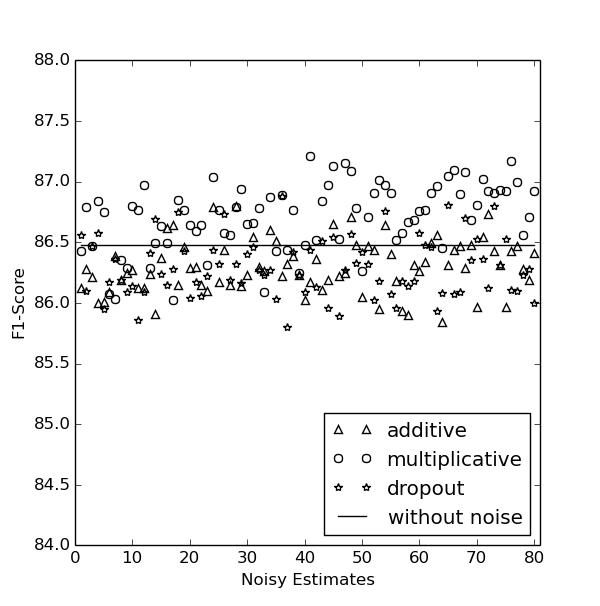}
  
  \caption{$F_1$ scores of noisy models. Each data point gives the
    $F_1$ accuracy of a single model on the development set, based on
    the legend. The $x$-axis enumerates the models (80 in total for each noise scheme).}
  \label{noisy-f1}
\end{figure}

\paragraph{Preliminary experiments} We first experiment with the
number of latent states for the clustering algorithm without
perturbations. We use $k=100$ for the SVD step. Whenever we need to cluster
a set of points, we run the
$k$-means algorithm $10$ times with random restarts and choose the
clustering result with the lowest objective value. On section 22, the
clustering algorithm achieves the following results ($F_1$ measure):
$m=8$: 84.30\%, $m=16$: 85.98\%, $m=24$: 86.48\%, $m=32$: 85.84\%,
$m=36$: 86.05\%, $m=40$: 85.43\%.  As we increase the number of
states, performance improves, but plateaus at $m=24$. For the rest of
our experiments, both with the spectral algorithm of
\newcite{cohenacl2012} and the clustering algorithm presented in this
paper, we use $m=24$.

\paragraph{Compact models} One of the advantage of the clustering
algorithm is that it leads to much more compact models. The number of
nonzero parameters with $m=24$ for the clustering algorithm is
approximately 97K, while the spectral algorithms lead to a
significantly larger number of nonzero parameters with the same number
of latent states: approximately 54 million.

\begin{table}[ht!]
  \begin{center}
    \begin{tabular}{|l|l|c|}
      \hline
      & Method & $F_1$ \\
      \hline
      \multirow{3}{*}{\rotatebox{90}{Best}}  & Spectral (unsmoothed) &  89.21 \\
      & Spectral (smoothed) & 88.87 \\
      & Clustering & 89.25 \\
      \hline
      \multirow{3}{*}{\rotatebox{90}{Hier}} & Spectral (unsmoothed) & 89.09 \\ 
      & Spectral (smoothed) & 89.06 \\
      & Clustering & \textbf{90.18} \\
      \hline
    \end{tabular}
  \end{center}
  \caption{Results on section 23 (English). The first three results
    (Best) are taken with the best model in each corresponding block
    in Table~\ref{table:results-dev}.  The last three results (Hier)
    use a hierarchy of the above tree combination methods in each
    block.  It combines all MaxEnt results using the maximal tree
    coverage (see text).\label{table:results-test}}
\end{table}

\paragraph{Oracle experiments} To what extent do we get a diverse
set of solutions from the different models we estimate? This question can be answered
by testing the oracle accuracy in the different settings. For each
type of noising scheme, we generated 80 models, 20 for each $\sigma
\in \{ 0.05, 0.1, 0.15, 0.2 \}$. Each noisy model by itself lags
behind the best model (see Figure \ref{noisy-f1}). However, when
choosing the best tree among these models, the additively-noised
models get an oracle accuracy of 95.91\% on section 22; the
multiplicatively-noised models get an oracle accuracy of 95.81\%; and
the dropout-noised models get an oracle accuracy of 96.03\%. Finally
all models combined get an oracle accuracy of 96.67\%. We found out
that these oracle scores are comparable to the one
\newcite{charniak-05} report. 


We also tested our oracle results, comparing the spectral algorithm of
\newcite{cohen-13b} to the clustering algorithm. We generated 20
models for each type of noising scheme, 5 for each $\sigma \in \{
0.05, 0.1, 0.15, 0.2 \}$) for the spectral algorithm.\footnote{There
  are two reasons we use a smaller number of models with the spectral algorithm:
  (a) models are not compact (see text) and (b) as such, parsing takes
  comparatively longer. However, in the above comparison, we use 20 models
for the clustering algorithm as well.} Surprisingly, even though the spectral models
were smoothed, their oracle accuracy was lower than the accuracy of
the clustering algorithm: 92.81\% vs. 95.73\%.\footnote{Oracle scores
  for the clustering algorithm: 95.73\% (20 models for each noising
  scheme) and 96.67\% (80 models for each noising scheme).} This
reinforces two ideas: (i) that noising acts as a regularizer, and has
a similar role to backoff smoothing, as we see below; and (ii) the
noisy estimation for the clustering algorithm produces a more diverse set of
parses than that produced with the spectral algorithm.

It is also important to note that the high oracle accuracy is not just the
result of $k$-means not finding the global maximum for the clustering
objective. If we just run the clustering algorithms with 80 models as
before, without perturbing the features, the oracle accuracy is
95.82\%, which is lower than the oracle accuracy with the additive and
dropout perturbed models. To add to this, we see below that perturbing
the training set with the spectral algorithm of Cohen et al. improves the
accuracy of the spectral algorithm. Since the spectral algorithm of Cohen
et al. does not maximize any objective locally, it shows that the role
of the perturbations we use is important.

\begin{table*}[htb]
  
  \begin{center}
    {\small
      \begin{tabular}{|c||c|c|c||c|c|c||c|c|c|}
        \hline
        & \multicolumn{3}{c||}{Clustering} & \multicolumn{3}{c||}{Spectral (smoothing)} & \multicolumn{3}{c|}{Spectral (no smoothing)} \\
        & MaxTre & MaxMrg & MaxEnt & MaxTre & MaxMrg & MaxEnt & MaxTre & MaxMrg & MaxEnt \\
        \hline
        \setupAdd & 77.34 & 76.87 & 80.01 & 77.76 & 77.85 & 78.09 & 77.44 & 77.56 & 77.91 \\
        \setupMul & 77.80 & 77.80 & 80.34 & 77.80 & 77.76 & 78.89 & 77.62 & 77.85 & 78.94 \\
        \setupDropout & 77.37 & 77.17 & \textbf{80.94} & 77.94 & 78.06 & 79.02 & 77.97 & 78.17 & 79.18 \\
        \setupAll & 77.71 & 77.51 & 80.86 & 78.04 & 77.89 & \textbf{79.46} & 77.73 & 77.91 & \textbf{79.66} \\
        \hline
        No noise & \multicolumn{3}{c||}{75.04} & \multicolumn{3}{c||}{77.71} & \multicolumn{3}{c|}{77.07} \\
        \hline
      \end{tabular}
    }
  \end{center}
  \caption{\small Results on the development set for German. See
    Table~\ref{table:results-dev} for interpretation of MaxTre,
    MaxMrg, MaxEnt and \setupAdd, \setupMul $\,$ and \setupDropout.
    The number of models used in the first three rows for the
    clustering algorithm is 80: 20 for each $\sigma \in \{ 0.05, 0.1,
    0.15, 0.2 \}$.  For the spectral algorithm, it is 20, 5 for each
    $\sigma$.  The number of latent states is $m=8$. For \setupAll, we
    use all models combined from the first three rows. The best figure
    in each algorithm block is in boldface.
    \label{table:results-dev-german}}
  \vspace{-0.15cm}
\end{table*}

\paragraph{Results} Results on the development set are given in
Table~\ref{table:results-dev} with our three decoding methods. We
present the results from three algorithms: the clustering algorithm
and the spectral algorithms (smoothed and
unsmoothed).\footnote{\newcite{cohen-13b} propose two variants of
  spectral estimation for L-PCFGs: smoothed and unsmoothed. The
  smoothed model uses a simple backedoff smoothing method which leads
  to significant improvements over the unsmoothed one. Here we compare
  our clustering algorithm against both of these models. However
  unless specified otherwise, the spectral algorithm of
  \newcite{cohen-13b} refers to their best model, i.e. the smoothed
  model.}


It seems that dropout noise for the spectral algorithm acts as a
regularizer, similarly to the backoff smoothing techniques that are
used in \newcite{cohen-13b}.  This is evident from the two spectral
algorithm blocks in Table~\ref{table:results-dev}, where dropout noise
does not substantially improve the smoothed spectral model (Cohen et
al. report accuracy of 88.53\% with smoothed spectral model for $m=24$
without noise) -- the accuracy is 88.64\%--88.71\%--89.47\%, but the
accuracy substantially improves for the unsmoothed spectral model,
where dropout brings an accuracy of 86.47\% up to 89.52\%.

All three blocks in Table~\ref{table:results-dev} demonstrate that decoding with
the MaxEnt reranker performs the best. Also it is interesting to note
that our results continue to improve when combining the output of
previous combination steps further.  The best result on section 22 is
achieved when we combine, using maximal tree coverage, all MaxEnt
outputs of the clustering algorithm (the first block in
Table~\ref{table:results-dev}). This yields a 90.68\% $F_1$
accuracy. This is also the best result we get on the test set (section
23), 90.18\%. See Table~\ref{table:results-test} for results on
section 23.


Our results are comparable to state-of-the-art results for
parsing. For example, \newcite{sagae2006parser},
\newcite{fossum2009combining} and \newcite{zhang2009k} report an
accuracy of 93.2\%-93.3\% using parsing recombination;
\newcite{shindo-12} report an accuracy of 92.4 $F_1$ using a Bayesian
tree substitution grammar; \newcite{petrov2010products} reports an
accuracy of 92.0\% using product of L-PCFGs; \newcite{charniak-05}
report accuracy of 91.4 using a discriminative reranking model;
\newcite{carreras08tag} report 91.1 $F_1$ accuracy for a
discriminative, perceptron-trained model; \newcite{petrov-07} report
an accuracy of 90.1 $F_1$. \newcite{collins-03} reports an accuracy of
88.2 $F_1$.


\subsection{Results for German}

For the German experiments, we used the NEGRA corpus \cite{skut97}. We
use the same setup as in \newcite{petrov2010products}, and use the first
18,602 sentences as a training set, the next 1,000 sentences as a
development set and the last 1,000 sentences as a test set.  This
corresponds to an 80\%-10\%-10\% split of the treebank. 

Our German experiments follow the same setting as in our English
experiments. For the clustering algorithm we generated 80 models, 20
for each $\sigma \in \{ 0.05, 0.1, 0.15, 0.2 \}$. For the spectral
algorithm, we generate 20 models, 5 for each $\sigma$.

For the reranking experiment, we had to modify the BLLIP parser
\cite{charniak-05} to use the head features from the German
treebank. We based our modifications on the documentation for the
NEGRA corpus (our modifications are based mostly on mapping of nonterminals to
coarse syntactic categories).

\paragraph{Preliminary experiments} For German, we also experiment
with the number of latent states. On the development set, we observe
that the $F_1$ measure is: 75.04\% for $m=8$, 73.44\% for $m=16$ and
70.84\% for $m=24$. For the rest of our experiments, we fix the number
of latent states at $m=8$. 



\begin{table}[htb]
\begin{center}
\begin{tabular}{|l|l|c|}
\hline
& Method & $F_1$ \\
\hline
\multirow{3}{*}{\rotatebox{90}{Best}}  & Spectral (unsmoothed) &  80.88\\
 & Spectral (smoothed) & 80.31 \\
 & Clustering &  81.94 \\
\hline
\multirow{3}{*}{\rotatebox{90}{Hier}} & Spectral (unsmoothed) & 80.64 \\ 
 & Spectral (smoothed) & 79.96 \\
& Clustering &  \textbf{83.38} \\
\hline
\end{tabular}
\end{center}
\caption{Results on the test set for the German data. The first three results (Best) are taken 
  with the best model in each corresponding block in Table~\ref{table:results-dev-german}. The last three results (Hier) use
  a hierarchy of the above tree combination methods. 
  \label{table:results-test-german}}
\vspace{-0.4cm}
\end{table}

\paragraph{Oracle experiments} The additively-noised models get an
oracle accuracy of 90.58\% on the development set; the
multiplicatively-noised models get an oracle accuracy of 90.47\%; and
the dropout-noised models get an oracle accuracy of 90.69\%. Finally
all models combined get an oracle accuracy of 92.38\%. 

We compared our oracle results to those given by the spectral
algorithm of \newcite{cohen-13b}.  With 20 models for each type of
noising scheme, all spectral models combined achieve an oracle
accuracy of 83.45\%. The clustering algorithm gets the oracle
score of 90.12\% when using the same number of models.

\paragraph{Results} Results on the development set and on the test set
are given in Table~\ref{table:results-dev-german} and
Table~\ref{table:results-test-german} respectively. 

Like English, in all three blocks in Table~\ref{table:results-dev-german},
decoding with the MaxEnt reranking performs the best. Our results
continue to improve when further combining the output of previous combination
steps. The best result of 82.04\% on the development set is
achieved when we combine, using maximal tree coverage, all MaxEnt
outputs of the clustering algorithm (the first block in
Table~\ref{table:results-dev-german}). This also leads to the best
result of 83.38\% on the test set. See
Table~\ref{table:results-test-german} for results on the test set.

Our results are comparable to state-of-the-art results for German
parsing. For example, \newcite{petrov2010products} reports an accuracy
of 84.5\% using product of L-PCFGs;
\newcite{petrov-07} report an accuracy of 80.1 $F_1$; and
\newcite{dubey2005} reports an accuracy of 76.3 $F_1$.

\section{Discussion}

From a theoretical point of view, one of the great advantages of
spectral learning techniques for latent-variable models is that they
yield consistent parameter estimates. Our clustering algorithm for
L-PCFG estimation breaks this, but there is a work-around to obtain an
algorithm which would be statistically consistent.

The main reason that our algorithm is not a consistent estimator is
that it relies on $k$-means clustering, which maximizes a non-convex
objective using hard clustering steps.  The $k$-means algorithm can be
viewed as ``hard EM'' for a Gaussian mixture model (GMM), where each
latent state is associated with one of the mixture components in the
GMM. This means that instead of following up with $k$-means, we could
have identified the parameters and the posteriors for a GMM, where the
observations correspond to the vectors that we cluster. There are now
algorithms, some of which are spectral, that aim to solve this
estimation problem with theoretical guarantees
\cite{vempala2004spectral,kannan2005spectral,moitra2010settling}.

With theoretical guarantees on the correctness of the posteriors from
this step, the subsequent use of maximum likelihood estimation step
could yield consistent parameter estimates. The consistency guarantees
will largely depend on the amount of information that exists in the
base feature functions about the latent states according to the L-PCFG
model.

\section{Conclusion}
\label{section:conclusion}

We presented a novel estimation algorithm for latent-variable PCFGs.
This algorithm is based on clustering of continuous tree representations,
and it also leads to sparse grammar estimates and compact
models. We also showed how to get a diverse set of parse tree predictions
with this algorithm and also older spectral algorithms.
Each prediction in the set is made by training an L-PCFG model
after perturbing the underlying features that estimation algorithm uses from the training data. We
showed that such a diverse set of predictions can be used to improve the parsing
accuracy of English and German.

\section*{Acknowledgements}

The authors would like to thank David McClosky for his help
with running the BLLIP parser and the three anonymous reviewers for
their helpful comments. This research was supported by an EPSRC
grant (EP/L02411X/1).


\bibliographystyle{acl}
\bibliography{mjc,mvl,cca,parsing,bib,nlp}

\end{document}